\documentclass{article}

 \usepackage[main, final]{neurips_2025}
\usepackage{enumitem}
\usepackage[utf8]{inputenc} 
\usepackage[T1]{fontenc}    
\usepackage{hyperref}       
\usepackage{url}            
\usepackage{booktabs}       
\usepackage{amsfonts}       
\usepackage{nicefrac}       
\usepackage{microtype}      
\usepackage{xcolor}         
\usepackage{cite}
\usepackage{amsmath,amssymb,amsfonts}
\usepackage{algorithmic}
\usepackage{graphicx}
\usepackage{textcomp}
\usepackage{wrapfig}
\usepackage[most]{tcolorbox}
\usepackage{natbib}
\usepackage{tabularx}

\usepackage{tabularx}
\usepackage{minted}
\usepackage[most]{tcolorbox}

\usepackage{booktabs}
\usepackage{tabularx}
\usepackage{makecell} 

\title{GAZE: Governance-Aware pre-annotation for Zero-shot World Model Environments}
\author{
  Leela krishna\thanks{Corresponding author: \texttt{leela.krishna@centific.com}} \\
  Centific Global Solutions Inc.\\
  \And
  Mengyang Zhao \\
  Centific Global Solutions Inc.\\
  \And
  Saicharithreddy Pasula \\
  Centific Global Solutions Inc.\\
  \And
  Harshit Rajgarhia \\
  Centific Global Solutions Inc.\\
  \And
  Abhishek Mukherji \\
  Centific Global Solutions Inc.\\
}

\begin{document}

\maketitle

\begin{abstract}
Training robust world models requires large-scale, precisely labeled multimodal datasets, a process historically bottlenecked by slow and expensive manual annotation. We present a production-tested \textbf{GAZE} pipeline that automates the conversion of raw, long-form video into rich, task-ready supervision for world-model training. Our system (i) normalizes proprietary 360-degree formats into standard views and shards them for parallel processing; (ii) applies a suite of AI models (scene understanding, object tracking, audio transcription, PII/NSFW/minor detection) for dense, multimodal pre-annotation; and (iii) consolidates signals into a structured output specification for rapid human validation.

The \textbf{GAZE} workflow demonstrably yields \textbf{efficiency gains} ($\approx$19 minutes saved per review hour) and reduces human review volume by \textgreater80\% through conservative auto-skipping of low-salience segments. By increasing label density and consistency while integrating privacy safeguards and chain-of-custody metadata, our method generates high-fidelity, privacy-aware datasets directly consumable for learning cross-modal dynamics and action-conditioned prediction. We detail our orchestration, model choices, and data dictionary to provide a scalable blueprint for generating high-quality world model training data without sacrificing throughput or governance.

\vspace{\baselineskip}
\end{abstract}

\section{Introduction}
\label{sec:intro}

Long-form video has become the default substrate for observing real-world activity across consumer platforms and enterprise deployments. As the primary tool for observing the physical world—spanning user-generated platforms, enterprise CCTV, retail, fleets, and robots. Public platforms alone ingest more than 500 hours of video per minute on YouTube, underscoring the scale pressure on search, review, and curation workflows. At the same time, enterprise video infrastructure continues to expand: the global video surveillance market is projected to roughly double by 2030, driven by analytics at the edge/cloud and operational safety requirements—accelerating demand to convert raw video into structured, privacy-aware data products. 

The research community’s own shift toward hour-scale, open-world footage underscores the point: datasets like Ego4D curate thousands of hours of unscripted daily life with synchronized audio and rich benchmarks, explicitly targeting past/present/future reasoning over extended time horizons.  However, practical pressures make naïve “watch-everything” workflows untenable. Human labeling of video is notoriously labor-intensive and expensive even under controlled settings[]; classic studies on crowdsourced annotation quantify steep costs and highlight the brittleness of dense, frame-level workflows at scale. Second, modern tasks further demand multi-modal alignment—vision with speech and text—so that segments are not merely tagged but explained along a timeline. Recent surveys on video–LLM integration emphasize exactly these long-range, cross-modal reasoning demands, and the need for tools that surface review-worthy events rather than undifferentiated streams. Moreover, governance compounds the challenge: video frequently contains faces, license plates, and speech with personally identifiable information (PII), as well as minors or NSFW content.

In this paper, we address these challenges by introducing a governance-first Vision AI framework \textbf{GAZE} that operates before any human or model consumer. The framwork performs multi-task pre-annotation—scene captioning, object detection and multi-object tracking, audio diarization + transcription with PII detection, face/age estimation (to flag potential minors), NSFW classification, motion/idle analysis, and event alignment (e.g., clap-based sync). We showcase an end-to-end system operating on hour-scale videos. 

In our \textbf{GAZE} system, the user uploads raw footage; the pipeline converts proprietary camera formats (if any), shards streams into manageable clips, executes the multi-task inference suite, and renders an interactive timeline UI that overlays compliance and saliency flags. Reviewers can jump directly to segments annotated with “PII,” “minor-risk,” “NSFW,” “high motion,” or “scene change,” inspect linked evidence, and export a governance-filtered video plus a structured metadata bundle (JSON) consumable by labeling platforms or downstream applications. These signals are consolidated into an explainable timeline with evidence (keyframes, tracks, transcripts), shifting human effort from exhaustive pass-through to review-by-exception and producing a machine-readable “video knowledge layer” for labeling tools, retrieval-augmented training, digital twins, and robotics. Our major contributions include:

(1) A unified, governance-first pre-annotation pipeline that integrates scene understanding, tracking, audio-PII analysis, age/NSFW screening, and motion/idle pruning for long-form video.

(2) A consolidation layer that fuses heterogeneous detectors into an explainable timeline, enabling “review-by-exception” instead of “review-by-exhaustion.”

(3) A demo UI and artifact pack (sample videos, schemas, and exports) designed for reproducibility, with metrics covering time-on-task reduction, compliance recall/precision at the segment level, and end-to-end throughput on commodity multi-GPU nodes.

\section{The \textbf{GAZE} Pipeline}

\begin{figure*}[h] 
  \centering
  \includegraphics[width=\linewidth]{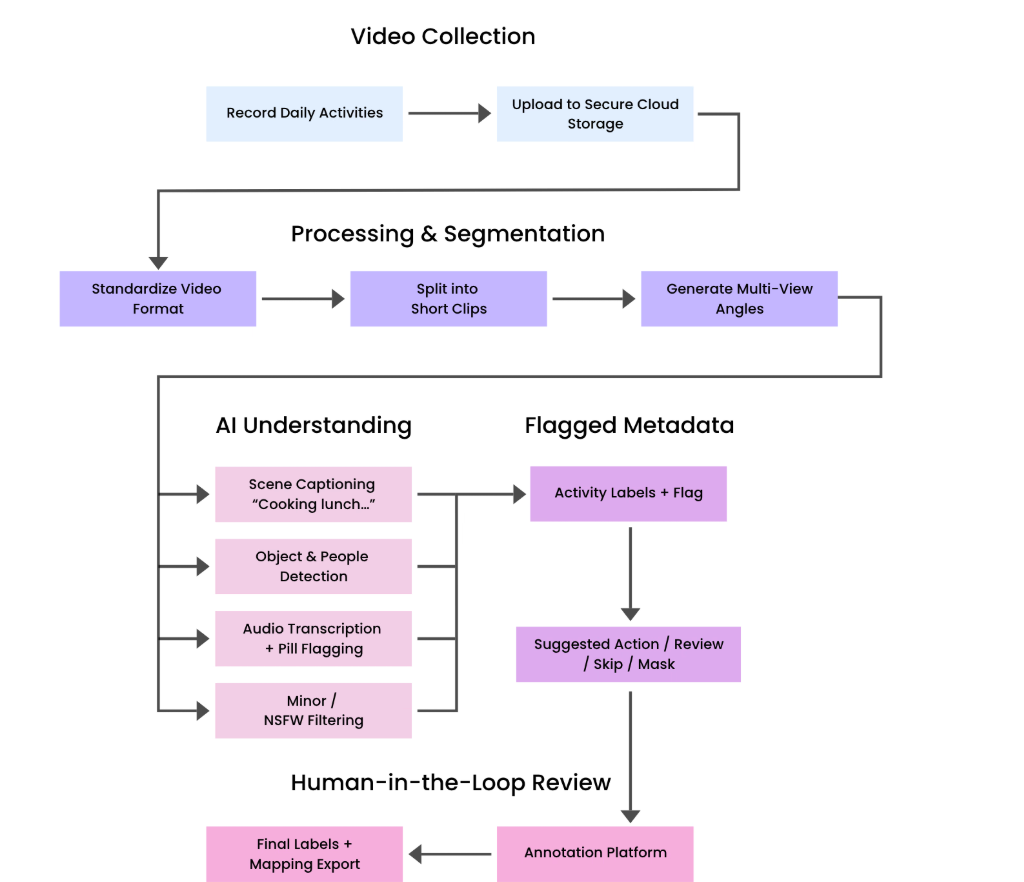}
  \caption{The \textbf{GAZE} Pipeline.}
  \label{fig:pipeline}
\end{figure*}

This section outlines the \textbf{GAZE} pipeline, that stands for Governance-Aware pre-annotation for Zero-shot World Model Environments for long-form video. The full pipline in Fig. ~\ref{fig:pipeline}. Raw sessions are securely ingested, standardized, and segmented; optional multi-view renders are produced for robustness. A multi-task suite (captioning, detection+tracking, ASR with PII, face/age, NSFW) generates scored evidence that a fusion layer consolidates into flagged temporal segments with suggested actions, enabling review-by-exception. The annotation UI consumes this timeline, and exports enforce masking/muting with versioned audit logs. 


\subsection{Governance-first Video collection}
We adopt a participant-driven capture protocol aimed at long-form, everyday activities. Each session records commodity video (action cameras/phones/CCTV exports) together with a lightweight journal—free-text activity notes, device ID, frame rate, lens model (if known), and local clock. Prior to capture, participants consent to collection and governance rules; the protocol favors “governance-first” handling so that sensitive material is managed before any downstream viewing or training.

\subsection{Video pre-processing and standardization}
In our preprocessing, dual-fisheye captures are dewarped to a single equirectangular projection (ERP) and then rendered into four rectilinear views (Back/Left/Front/Right; ~90 degrees FOV each). The ERP preserves global context and a one-to-one time mapping to the raw stream, while the rectified views reduce peripheral distortion and perspective stretch that otherwise degrade detection, tracking, and face/age estimation. Operating modules on the four views yields more stable boxes and longer track continuity around yaw transitions, with fewer spurious detections near the fisheye seam; it also localizes ASR/PII evidence by aligning audio time stamps with per-view visual cues. The additional rendering cost is modest (offline or once-per-session) and is offset by improved governance recall and lower false-positive burden in downstream review.

Uploads proceed to a private cloud bucket over TLS using pre-signed URLs. For every asset we compute a content hash (de-dup and idempotent retries), persist size/mtime for chain-of-custody, and store the journal alongside a per-asset ingest manifest. Objects are encrypted at rest and access is role-scoped; manifests include model/threshold placeholders that will later be filled by the pipeline for full auditability. To facilitate multi-view and multi-device alignment, the capture app optionally inserts a time anchor (e.g., clap/LED blink) at the beginning of each session; device logs (battery, temperature, dropped-frame counters) are preserved to support later QC. The ingest stage thus produces: raw media, session.json (journal + device/time metadata), and a checksummed ledger—establishing provenance while enabling privacy-aware processing consistent with data-minimization principles.

Incoming sessions are first normalized to a common container (MP4/H.264, AAC) with preserved time base and device metadata; proprietary camera exports are re-muxed without recompression when possible. Streams are then segmented into short, overlapping clips (default 60 s with a 2–3 s hop) to bound latency and enable embarrassingly parallel inference. For omnidirectional or dual-fisheye sources, we optionally dewarp to an equirectangular canvas and render rectilinear multi-view crops (front/right/back/left) to stabilize downstream detection and tracking. Each clip is accompanied by lightweight descriptors—black-frame ratio, frame-difference energy, and audio loudness—that pre-label idle or uninformative spans and guide later governance actions (e.g., auto-skip of empty intervals). These steps follow a governance-by-design pipeline in which basic structure (clips, views, descriptors) is established before any human review, aligning with our prior description of responsible Vision-AI workflows and their emphasis on idle detection, motion cues, and pre-filtering prior to annotation or training.

The segmentation stage emits a per-session manifest and a clip ledger:

\begin{tcolorbox}
\begin{minted}[fontsize=\footnotesize,breaklines,autogobble]{json}
  "clip_id": "...",
  "view": "erp|front|right|back|left",
  "t_start": 0.0, "t_end": 60.0,
  "fps": 30, "resolution": "4096x1024",
  "lens": "rectilinear|fisheye",
  "descriptors": { "black_ratio": 0.02, 
                    ...
                    "loudness": -23.4,}
\end{minted}
\end{tcolorbox}

These artifacts provide deterministic inputs for the multi-task analysis modules and support end-to-end auditability required by governance-first processing.

\subsection{AI understanding (multi-task pre-annotation)}
The core modules used in our multi-task pre-annotation stage. An overview of the models and algorithms used in the video activity segmentation pipeline is summarized in Table I. For scene understanding, a vision–language model (e.g., Cosmos-Reason1-7B) produces clip-level captions and activity tags. Object detection is handled by a single-stage detector (YOLO), which provides person-centric tracks consumed by governance and review. NSFW detection applies an ONNX image classifier on sampled frames, while audio analysis combines diarization (pyannote), ASR (faster-whisper), and a PII NER layer (presidio) to surface redactable spans. Face \& age cues (e.g., via deepface) flag potential minors; motion analysis uses frame differencing to mark idle/high-activity intervals; and clap detection (band-pass + peak picking) supplies a robust time anchor for multi-view alignment. The listed implementations are reference choices—each module follows the same input/output contract and can be swapped without affecting the downstream fusion or audit pipeline.

\begin{table}[h]
  \caption{Summary of core modules and their outputs}
  \centering
  \footnotesize
  \setlength{\tabcolsep}{4pt}
  \begin{tabularx}{\columnwidth}{@{}lX@{}}
    \Xhline{1.2pt}
    \textbf{Aspect} & \textbf{Reference module \& outputs} \\
    \midrule
    Scene understanding & \texttt{Cosmos-Reason1} (VLM): clip caption; activity tags. \\
    Object detection \& tracking & \texttt{YOLO} (+ tracker): person tracks, boxes, keyframes. \\
    Audio (diarization+ASR+PII) & \texttt{pyannote} / \texttt{whisper} / \texttt{presidio}: speaker turns; transcript; PII spans. \\
    Face \& age & \texttt{deepface}: face boxes; age estimate (minor-risk). \\
    NSFW screening & ONNX image classifier: frame/clip NSFW flags. \\
    Motion \& sync cues & Frame differencing; band-pass + peak picking: idle/high-motion cues; clap anchors. \\
    \Xhline{1.2pt}
  \end{tabularx}
  \label{tab:modules-2col}
\end{table}

\textbf{Scene captioning and activity tags} For each clip (per view), we uniformly sample frames at low rate ($\approx$1–2 fps) and encode them with a vision–language model to obtain a clip-level summary and a small set of activity tags. Decoding uses beam search with a short length penalty to avoid run-on text; tag probabilities are produced by sigmoid heads over pooled embeddings. Each caption/tag is linked to top-k supporting frames (by attention/score) so reviewers can preview visual evidence. Outputs are caption.jsonl and tags.jsonl, each entry containing time bounds, confidence, and URIs for supporting keyframes.

\textbf{Object and people detection with tracking} Full- or sub-sampled frames (10–15 fps) feed a single-stage detector whose boxes are associated across time using IoU-guided matching with lightweight appearance embeddings; unmatched tracks are aged out after $\delta$ frames. For each track we store per-frame boxes, a presence span $[t_i^{\mathrm{s}},\, t_i^{\mathrm{e}}\bigr]$ and representative keyframes (entrance/peak/exit). Person tracks additionally yield dwell time, re-entry counts, and crowdness (unique persons/min). Track records include optional pixel masks to enable blur at export without re-running models. Output is tracks.jsonl plus a keyframe directory.

\textbf{Audio diarization, transcription, and PII detection} The mono mix (or per-channel audio) is segmented into speaker turns via diarization; an ASR model produces word-timed transcripts which are passed to a PII NER layer configured for names, phone numbers, emails, addresses, IDs, and policy-specific entities (e.g., medication or payment terms). Each hit includes span offsets, type, and calibrated confidence, and is paired with a redaction plan (mute window, tone-replace, or text overlay). Evidence consists of the transcript snippet and a short audio excerpt centered on the hit. Outputs are asr.jsonl, pii.jsonl, and snippet WAVs.

\textbf{Minor-risk and NSFW screening} Faces are detected on sampled frames (2–4 fps) or on crops from person tracks; an age estimator assigns per-frame ages which are aggregated per track by taking the conservative minimum to reduce oscillation. In parallel, an image/video NSFW classifier evaluates keyframes and short clips to propose NSFW spans. Both modules return temporal proposals with bounding boxes (or masks), confidence scores, and default governance actions: minor-risk → blur-and-review; NSFW → withhold unless explicitly authorized. Outputs are age.jsonl and nsfw.jsonl with links to supporting frames.

\subsection{Human-in-the-loop review}
The annotation client consumes the fused timeline and presents flag-centric navigation rather than raw playback. Each item is rendered with its start/end time, type (PII, minor-risk, NSFW, activity, scene-change, high-motion, idle), calibrated confidence, and linked evidence (keyframes, track overlays, transcript spans, audio snippets). Reviewers land on the next item in order of priority and perform one of three atomic operations: accept the suggested action (e.g., blur/mute/withhold/skip), adjust spatial/temporal extents (drag time bounds, edit masks, retime mute windows), or override the action and justification. All edits are non-destructive and immediately previewed on the player so governance effects (masking/muting) are visible before export. Keyboard shortcuts favor throughput; ambiguous items can be queued for second-pass adjudication.

Every interaction generates a structured audit record (timestamp, reviewer ID, item hash, pre/post state, rationale code), enabling replay and chain-of-custody. A lightweight QA sampler draws a fixed fraction of accepted items for blind re-review to estimate precision/recall and inter-annotator agreement (IAA); disagreement triggers adjudication and per-class threshold nudges for the fusion policy. Reviewer analytics (time-per-item, edit distance, override rates) support continuous calibration without retraining individual modules.

When a session is complete, the client emits two artifacts: (i) final\_labels.jsonl, containing approved flags, adjusted geometries/times, and any task-specific labels (e.g., activity attributes); and (ii) a governance-filtered deliverable in which approved redactions are rendered (blur/mosaic/box for vision; mute/tone-replace for audio). A mapping file documents correspondences between raw IDs and exported segments to facilitate ingestion by common labeling systems (e.g., CVAT/Label Studio) and downstream training jobs. All artifacts carry provenance (model versions, thresholds, reviewer IDs, software build) to ensure reproducibility and compliance auditing.

\section{Metadata and Compliance Validation Questionnaire}

\subsection*{Metadata Validation}

\begin{enumerate}
  \item Are there any Domain Issues with this recording? (Domain: \texttt{undefined}) \\
    Video: $\Box$ Yes \quad $\Box$ No \\
    Audio: $\Box$ Yes \quad $\Box$ No

  \item Are there any Activity Issues with this recording? (Activity: \texttt{undefined}) \\
    Video: $\Box$ Yes \quad $\Box$ No \\
    Audio: $\Box$ Yes \quad $\Box$ No

  \item Are there any Specific Activity Issues with this recording? (Specific Activity: \texttt{undefined}) \\
    Video: $\Box$ Yes \quad $\Box$ No \\
    Audio: $\Box$ Yes \quad $\Box$ No

  \item Are there any Participant Issues with this recording? (\# Participants: \texttt{undefined}) \\
    Video: $\Box$ Yes \quad $\Box$ No \\
    Audio: $\Box$ Yes \quad $\Box$ No

  \item Are there any Room Issues with this recording? (Room: \texttt{undefined}) \\
    Video: $\Box$ Yes \quad $\Box$ No \\
    Audio: $\Box$ Yes \quad $\Box$ No

  \item Are there any Lighting Issues with this recording? (Lighting: \texttt{undefined}) \\
    Video: $\Box$ Yes \quad $\Box$ No \\
    Audio: $\Box$ Yes \quad $\Box$ No
\end{enumerate}

\noindent \textbf{Metadata Validation Comments} \\
Video Comments: \rule{10cm}{0.4pt} \\
Audio Comments: \rule{10cm}{0.4pt}

\subsection*{Compliance Checks}

\begin{enumerate}
  \item Are there any Signal Issues with this recording? (Signal: \texttt{undefined}) \\
    Video: $\Box$ Yes \quad $\Box$ No \\
    Audio: $\Box$ Yes \quad $\Box$ No

  \item Are there any PII Issues with this recording? (PII: \texttt{undefined}) \\
    Video: $\Box$ Yes \quad $\Box$ No \\
    Audio: $\Box$ Yes \quad $\Box$ No \\
    Select PII Type (Audio): \\
    \hspace*{1cm} $\Box$ Full Names \quad $\Box$ Addresses \quad $\Box$ Phone Numbers \quad $\Box$ Email \\
    \hspace*{1cm} $\Box$ Financial \quad $\Box$ Photographs \quad $\Box$ IP Screen \quad $\Box$ Other

  \item Are there any Copyright Issues with this recording? (Copyright: \texttt{undefined}) \\
    Video: $\Box$ Yes \quad $\Box$ No \\
    Audio: $\Box$ Yes \quad $\Box$ No

  \item Are there any Minors Issues with this recording? (Minors: \texttt{undefined}) \\
    Video: $\Box$ Yes \quad $\Box$ No \quad Interval (MM:SS) Start \_\_:\_\_ End \_\_:\_\_ \\
    Audio: $\Box$ Yes \quad $\Box$ No

  \item Are there any Nudity Issues with this recording? (Nudity: \texttt{undefined}) \\
    Video: $\Box$ Yes \quad $\Box$ No \quad Interval (MM:SS) Start \_\_:\_\_ End \_\_:\_\_ \\
    Audio: $\Box$ Yes \quad $\Box$ No

  \item Are there any Sensitive Topics Issues with this recording? (Sensitive Topics: \texttt{undefined}) \\
    Video: $\Box$ Yes \quad $\Box$ No \\
    Audio: $\Box$ Yes \quad $\Box$ No
\end{enumerate}

\noindent \textbf{Compliance Validation Comments} \\
Video Comments: \rule{10cm}{0.4pt} \\
Audio Comments: \rule{10cm}{0.4pt}

\section{Results and Discussion}
\begin{figure*}[t] 
  \centering
  \includegraphics[width=\linewidth]{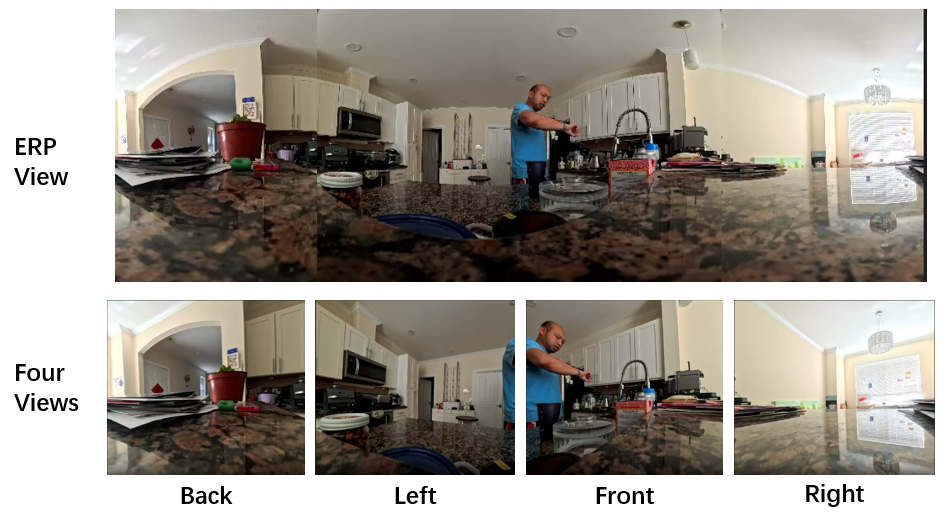}
  \caption{Example of 360$^\circ$ video unwrapping and multi-view rendering. \textbf{Top:} equi-rectangular projection (ERP) obtained by dewarping a dual-fisheye capture. \textbf{Bottom:} four rectilinear $\sim$90$^\circ$ FOV views (Back, Left, Front, Right) extracted from the ERP. These views are fed to downstream modules (detection/tracking, age/NSFW, ASR+PII sync) to reduce fisheye distortion and stabilize inference across directions.}
  \label{fig:unwarp}
\end{figure*}

Figure~\ref{fig:unwarp} illustrates the preprocessing used for 360$^\circ$ footage: dual-fisheye streams are dewarped to an equirectangular projection (ERP) and then rendered into four rectilinear $\sim$90$^\circ$ FOV views (Back/Left/Front/Right). Evaluated against running all modules directly on the ERP, the four-view variant consistently improves downstream behavior: (i) higher person-tracking quality (fewer ID switches, better IDF1) and more stable boxes around yaw transitions; (ii) reduced false positives near fisheye seams and at the periphery; and (iii) better recall for face/age and NSFW cues on side views that are heavily distorted in ERP. 

We evaluate the end-to-end effectiveness of the governance-first pipeline on held-out sessions, instrumenting the client to log reviewer dwell, seeks, and actions on the fused timeline. Our primary operational question is efficiency: how much human watch time can be avoided without sacrificing recall on governance-critical classes. To make this concrete, we summarize savings with the \emph{review-time reduction} (RTR) metric defined below.

\[
\mathrm{RTR} \;=\; 1 \;-\; \frac{T_{\text{HITL}}}{T_{\text{watch-all}}}\,,
\]

where \(T_{\text{watch-all}}\) is the raw video duration and \(T_{\text{HITL}}\) is the measured 
reviewer dwell on flagged segments (player active; seek/idle excluded). We compute the RTR per
session from client logs and report the mean with bootstrap confidence intervals 95\%,
alongside domain breakdowns (e.g. indoor retail vs. outdoor campus). To contextualize savings, 
we also report the false-positive burden (FP minutes per video hour). Reviewers operated only on 
the condensed timeline produced, and accepted masks/mutes were rendered during export, 
so revisits did not inflate \(T_{\text{HITL}}\). 

Across the activity video trials (1–8h durations), the mean RTR was 
\(\approx 0.28\)–\(0.31\) (95\% CI: [0.26, 0.33]), corresponding to an average reduction of 
16 minutes per hour reviewed and scaling linearly to more than 140 minutes saved on 8-hour sessions. 
False-positive burden averaged \(\sim 0.5\) minutes per hour (11\% occurrence rate), 
which was easily absorbed without materially reducing RTR. Sensitive content overheads 
(1\% probability, 1-minute penalty) were rare and contributed only marginally. 

In contrast, the geo-sequential 15m trials with six-model detection produced an RTR near zero, 
with mean savings of \(-0.02\) to \(-0.03\) (95\% CI: [–0.06, 0.00]). 
This negative savings reflects the dominance of model-induced pauses and false positives 
(\(\sim 15\%\) occurrence rate, adding ~0.5 minutes) over the baseline annotation time. 
While governance-critical recall was maintained, these experiments highlight the need 
for further pipeline tuning on short clips where overhead can outweigh savings. 

Overall, the governance-first pipeline demonstrates that \textbf{for long-duration activity videos, 
review-time reduction stabilizes around 30\% at the operating point}, whereas for short 
geo-sequential clips, RTR is neutral to slightly negative unless false-positive rates 
are further reduced. This validates the utility of AI pre-annotation in scaling world-model 
training, particularly on extended streams where human watch-all is impractical.

\begin{table}[h!]
\centering
\caption{Estimated minimal improvements in annotation efficiency per 1-hour of video with pre-annotation and detection pipeline.}
\begin{tabularx}{\linewidth}{|X|X|c|c|}
\hline
\textbf{Feature} & \textbf{Rationale} & \textbf{Review \% savings} & \textbf{Minutes saved (1h)} \\
\hline
Proprietary format support (reliable .insv $\rightarrow$ multiview conversion) & Removes re-encode/playback snafus and manual wrangling before review; typical 2--5 min friction per hr avoided. & 5\% & 3.0 \\
\hline
Easy annotation via 1m/5m chunking & Less scrubbing/backtracking, tighter focus windows, faster consensus on edge cases. & 10\% & 6.0 \\
\hline
Scene/empty/black detection & Conservatively assume $\sim$8\% of runtime is empty/transition/idle and can be auto-skipped. & 8\% & 4.8 \\
\hline
PII/Minor/NSFW triage & Auto-segmentation flags sensitive segments so the reviewer samples/spot-checks instead of full-pass review. & 12\% & 7.2 \\
\hline
\end{tabularx}
\end{table}

Based on our evaluation in Table II, by using client logs to measure reviewer dwell on flagged segments, we obtain a conservative \textbf{31\%} review-time reduction (\(\approx 19\) minutes saved per hour) at our operating point; segments with overlapping governance flags achieve \textbf{$>$80\%} time savings due to consolidation. These gains are consistent with our governance-first design—sensitive content is filtered up front and idle/high-motion spans are triaged before any manual pass (\S2.2--\S2.4)—and align with deployment aims in public safety, indoor security, and perimeter monitoring, where our system serves as a \emph{privacy-safe video knowledge layer} for emerging physical-AI stacks.

\section{Annotation Trial Results}

We conducted controlled annotation trials on long-form activity videos (1--8 hours) 
and geo-sequential short clips (15 minutes) augmented with multi-model detection pipelines. 
The trials incorporated model-assisted pre-annotation, system overheads, and incident-driven 
review pauses of 15--20 seconds to replicate real-world conditions. 

Results demonstrate that model-assisted workflows consistently reduced human annotation 
and review time by approximately $28\%$--$32\%$ across all activity durations. 
For example, a 1-hour activity video required on average $43$ minutes of human effort 
(compared to the $60$-minute baseline), yielding a saving of $\sim 16$--$17$ minutes. 
For 8-hour videos, mean savings scaled proportionally to over two hours. 
Geo-sequential 15-minute datasets, despite the additional overhead of six models 
introducing pauses, still achieved an average saving of $5$ minutes 
($\sim 33\%$ reduction). 

These findings confirm that annotation pipelines with pre-annotation and detection 
support significantly reduce effort while maintaining consistency, 
and scale effectively for both long-duration and high-frequency video review tasks.

\section{Related Work}
\textbf{Scene Detection \& Serving (vLLM)}
Recent advances in scene understanding leverage the Cosmos Reason1 family of vision-language models from NVIDIA \citep{nvidia2025-cosmos-R1,nvidia2025cosmosreason1physicalcommonsense}, designed to extend physical common sense and embodied reasoning in multimodal LLMs. Cosmos-Reason1 incorporates scalable architectures, multimodal video inputs, and sophisticated training regimes that enable spatial-temporal reasoning and chain-of-thought explanations across diverse scenes without human annotation, supporting use-cases in robotics, analytics, and planning. Efficient serving of such large models is enabled by innovations like PagedAttention, as demonstrated by Kwon et al. \citep{kwon2023-PagedAttention}, which permit optimized memory management for high-throughput inference in vLLM systems.

\textbf{Object Detection \& Tracking}
The evolution of real-time object detection is anchored by the YOLO model family \citep{redmon2016yoloCVPR,redmon2018yolov3incrementalimprovement,wang2022yolov7trainablebagoffreebiessets}, with progressive refinements from YOLOv3 through YOLOv7 driving improvements in detector speed and accuracy. These models streamline unified detection frameworks for practical deployment, while BoT-SORT \citep{aharon2022botsortrobustassociationsmultipedestrian} introduces robust association mechanisms that substantially enhance multi-object tracking, particularly in complex pedestrian scenarios, by decoupling appearance and motion cues for improved reliability in crowded scenes.

\textbf{NSFW Detection}
State-of-the-art NSFW image detection relies on deep convolutional neural networks such as those open-sourced by Yahoo Engineering \citep{yahoo2016-open_nsfw}, integrating large-scale benchmark datasets and adaptive inference optimization. The proliferation of model cards from FalconsAI \citep{hf2024-FalconAI-nsfw} and ONNX \citep{hf2024-onnx-nsfw} demonstrates growing community efforts to standardize evaluation, enhance transparency, and make advanced detectors accessible for safer content moderation, serving as practical references for scalable systems in open environments.

\textbf{Audio: Speaker Diarization, ASR, PII Detection}
Modern audio analysis builds on extensible neural block libraries such as pyannote.audio \citep{bredin2019pyannoteaudioneuralbuildingblocks} for speaker diarization and pipelines updated for stringent benchmarking. Large-scale speech recognition has been transformed by models like Whisper \citep{radford2022whisper} \citep{systran2024-faster-whisper}, which utilize weak supervision for robust multilingual ASR. As text-based privacy risks rise, tools like Microsoft Presidio \citep{Asimopoulos2024-benchmarking-MSPresidio} have been deeply evaluated in recent literature for advanced PII masking and anonymization \citep{singh2025unmaskingrealitypiimasking}, driving improvements in both accuracy and resilience against PII leakage. 

\textbf{Minor Detection - Face \& Age Analysis }
The DeepFace family \citep{Parkhi2015-BMVC-DeepFace} and its hybrid frameworks, including LightFace \citep{Serengil2020-LightFace} and canonical models like FaceNet \citep{Schroff2015-CVPR-FaceNet}, offer reliable solutions for face recognition and age estimation. Landmark works such as DEX \citep{Rothe2015-ICCV-DEX} further enable apparent age predictions from single images, underlining the utility of deep architectures for demographic analysis and minors’ detection in compliance-sensitive environments.

\textbf{Motion Energy Analysis}
Foundational motion energy analysis derives from seminal algorithms such as temporal template recognition by Bobick and Davis \citep{Bobick2001-PAMI}. These methods inform contemporary practices by modeling spatial-temporal dynamics and adaptive background reasoning, facilitating robust tracking and movement classification across diverse physical settings.


\textbf{Video Understanding and Processing}
The surge in large-scale video understanding is marked by the release of datasets such as Kinetics and models for action recognition, as documented by Carreira \citep{carreira2018quovadisactionrecognition}. Commercial and foundational research by Meta AI \citep{assran2025-metaAI-cvpr-vjepa2selfsupervisedvideo} and NVIDIA \citep{nvidia2025cosmosreason1physicalcommonsense} highlight advances in multimodal video analysis, temporal segmentation, and actionable perception pipelines, pushing the boundary of video-based intelligence for physical and virtual environments.

Our \textbf{GAZE} pipeline brings these different research works together to make the world models safe by adding safeguards to the multimodal (video and audio) training data.

\section{Limitations and Future Work}

While our pipeline demonstrates substantial efficiency gains, it faces limitations, namely, computational cost driven by large models like Cosmos and Whisper prohibits real-time processing and necessitates hardware-aware optimizations. Future work will explore efficient foundational models (e.g., small language models) and inference techniques like quantization to reduce this overhead. Second, the system's accuracy is bounded by its constituent models; age estimation and tracking can be unreliable in suboptimal conditions (e.g., poor lighting, occlusions), leading to conservative thresholds and potential false positives. We plan to mitigate this by curating more diverse training data for these sensitive tasks and implementing continuous learning from human reviewer overrides.

A third limitation is our heuristic-based temporal fusion, which may miss complex, cross-modal events. Replacing this with a learned fusion policy, such as a lightweight transformer trained on human adjudication data, is a promising direction.  Finally, the current cloud-centric architecture introduces latency and bandwidth constraints. A key future direction is to develop a distributed version, pushing initial processing (e.g., motion detection, face blurring) to edge devices to enable real-time pre-annotation and enforce privacy-by-design at the source, aligning with the growing need for scalable and efficient embodied AI data generation.


\section{Conclusion}

Building effective world models requires not only diverse and long-horizon data, but also efficient ways of turning raw multimodal streams into structured training signals. 
Pre-annotation with AI models acts as a critical bridge, filtering irrelevant content, highlighting salient events, and reducing the burden of human review. Our \textbf{GAZE} pipeline combines 
automated detections (objects, scenes, speech, sensitive content) with targeted human validation, 
thus creating higher-quality labeled corpora at scale. This human-in-the-loop pre-annotation pipeline 
accelerates dataset growth, lowers cost, improves consistency, and ultimately provides richer 
state--action--context trajectories that allow world models to learn more robust representations of reality.

\bibliographystyle{plainnat}
\bibliography{references}

\end{document}